%% file: acl_latex.tex
\documentclass[11pt]{article}

\usepackage[final]{acl}

\usepackage{times}
\usepackage{latexsym}
\usepackage{amsmath}
\usepackage{amssymb} 
\usepackage{times}
\usepackage{latexsym}
\usepackage{pgfplots}
\pgfplotsset{compat=1.18}
\usepgfplotslibrary{groupplots}
\usepackage{subcaption}
\usepackage{newtxtext}
\usepackage{tikz}
\usetikzlibrary{arrows.meta,positioning}
\usepackage{amsthm} 
\usepackage{booktabs}

\usepackage{enumitem}
\usepackage{multirow}
\usepackage{makecell}
\usepackage[most]{tcolorbox}
\usepackage{pgfplots}
\usepgfplotslibrary{polar}
\pgfplotsset{compat=1.17}
\usepgfplotslibrary{fillbetween}
\usetikzlibrary{intersections} 
\usepackage[T1]{fontenc}

\usepackage[utf8]{inputenc}

\usepackage{microtype}

\usepackage{inconsolata}

\usepackage{graphicx}

\newtcolorbox{llmtext}{
  boxrule=0.5pt,
  colback=gray!10,
  arc=2pt,
  left=1pt, right=1pt, top=1pt, bottom=1pt,
  boxsep=2pt,
  before skip=2pt,
  after skip=2pt,
  enhanced,
  breakable,
  fontupper=\small, 
}

\title{Twin Worlds: Equivariance-Based Abstention \\for Evidence-Grounded Reasoning}

\author{
\textbf{
Vy Nguyen$^{1}$,
Ziqi Xu$^{1}$,
Jeffrey Chan$^{1}$,
Estrid He$^{1}$,
Feng Xia$^{1}$
} \\
\textbf{
Renqiang Luo$^{2}$,
Erik Cambria$^{3}$,
Xiuzhen Zhang$^{1*}$
} \\[1mm]
$^{1}$RMIT University, Australia \qquad
$^{2}$Jilin University, China \\
$^{3}$Nanyang Technological University, Singapore \\
\texttt{s3964786@student.rmit.edu.au, xiuzhen.zhang@rmit.edu.au} \\
{\small $^{*}$Corresponding author}
}

\begin{document}
\maketitle

\begin{abstract}
Knowledge-intensive reasoning requires Large Language Models (LLMs) to ground answers in provided evidence. When evidence is insufficient, it is desirable that models abstain rather than confidently generating unsupported answers. Existing abstention methods rely on uncertainty estimation or evidence sufficiency checks, but neither tests whether the reasoning process for generation, driven by the interaction of provided evidence and the model's internal memory parameters, is actually grounded in the evidence. A key contributing factor is that entity mentions in context activate memorised associations, causing models to generate plausible responses ungrounded in evidence. We propose \textbf{Twin Worlds (TW)}, a framework for improving reliability in knowledge-intensive reasoning through \textit{equivariance}-based abstention: unlike invariance, which requires outputs to remain unchanged, equivariance requires outputs to \textit{transform correspondingly} under entity substitutions. A model grounded in the evidence should produce answers that shift consistently when entities are substituted while their relations are preserved. TW constructs multiple worlds via typed substitutions of the original input that preserve relational structure while reducing parametric priors, and uses equivariance violations as an abstention signal. Across four benchmarks and three model backbones, TW identifies when answers are not reliably grounded in the provided evidence and outperforms uncertainty- and sufficiency-based baselines.\footnote{We release the source code at \url{https://github.com/xiuzhenzhang/Twin-Worlds}}\end{abstract}

\section{Introduction}

\begin{figure}[t]
    \includegraphics[width=1\columnwidth]{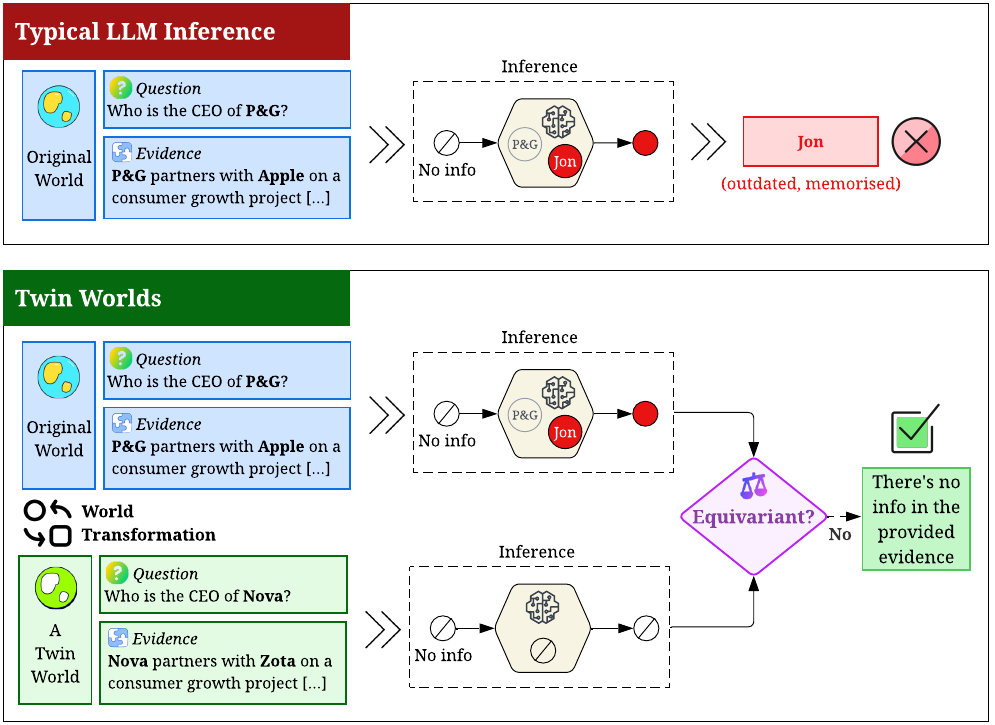}
    \caption{\textit{Top:} LLMs often answer confidently based on outdated parametric associations even when the provided evidence does not support a conclusion. \textit{Bottom:} Twin Worlds applies structure-preserving transformations to test whether answers are grounded in the evidence, enabling principled abstention decisions.}
    \label{fig:Example1}
\end{figure}

Large Language Models (LLMs) are increasingly used for knowledge-intensive reasoning with external evidence, where generation of answers requires combining a query with documents, retrieved passages, or tool outputs~\citep{Patrick2020RAGOriginal, yao2022react, qin2024toolllm}. A central failure in these settings is that models often continue to answer even when the provided evidence is insufficient, misleading, or internally conflicting~\citep{Park2025Mirage, Ming2025FaithEval}. Consider the question \textit{``Who is the CEO of P\&G?''} The accompanying evidence may fail to support it: passages may discuss P\&G's products or partnerships without identifying the CEO. Yet an LLM may still produce a plausible name consistent with reliance on memorised associations tied to the company mention (Fig.~\ref{fig:Example1}, top), rather than reasoning from the evidence~\citep{mallen-etal-2023-trust, Li2025Memorization}. The desired behaviour is to recognise when the available evidence does not justify an answer and to abstain.

Multi-evidence settings further complicate this problem, where contexts often contain distractor passages, partial support, or plausible cues for unsupported answers. With unfaithful evidence, irrelevant information can amplify parametric activation, causing confident answers from unrelated knowledge rather than abstention~\citep{Ming2025FaithEval}. Conversely, faithful evidence can still be overridden by conflicting parametric knowledge~\citep{Wu2024ClashevalTugofWar}, producing spurious uncertainty that triggers unnecessary abstention. Existing approaches ask either \textit{Is the model uncertain?}~\citep{Chen2024ControllingRisk, Wen2024CharacterizingPurtubation, Kim2025WhenToAbstain} or \textit{Is the evidence sufficient?}~\citep{Sun2025DivideAlign, Joren2025SufficientContext,RenLXZFL26}. However, a model may remain confident and self-consistent while following memorised associations that suggest a seemingly plausible but unjustified answer. We instead ask: \textit{Is the reasoning process driving generation actually grounded in the provided evidence rather than in memorised associations?}

We propose \textbf{Twin Worlds (TW)}, a framework for improving reliability in knowledge-intensive reasoning through \textit{equivariance}-based abstention: a model whose answers are grounded in the evidence should produce outputs that transform consistently under structure-preserving substitutions of entities. Many knowledge-intensive tasks, including open-domain question answering, fact verification, and multi-hop reasoning, are strongly entity-centric, with questions and evidence structured around named entities~\citep{Yang2018HotpotQA, Thorne2018Fever, mallen-etal-2023-trust,LiXRLZZRX26}. TW is therefore most naturally suited to settings with identifiable entity structure. LLMs encode substantial knowledge about these entities in their internal representations~\citep{hu-etal-2025-enabling, Sakata2025Entity, Morand2025EntityRepresentations,YangXLMCSZR26}, and entity mentions in context can activate this stored knowledge~\citep{Zhou_Xiang_Chen_Su_2024, farahani-johansson-2024-deciphering, ferrando2025do}, potentially driving generation from parametric memory. TW constructs multiple twin worlds, each defined by a typed bijective transformation of the original input that preserves relational structure while reducing parametric priors, and uses equivariance violations across these worlds as an abstention signal. 

When the evidence supports an answer, structure-preserving substitutions should induce corresponding answer transformations, yielding high equivariance. When evidence is incomplete, misleading, or inconsistent, equivariance may break, indicating that the answer is not reliably determined by the relational structure of the provided evidence and signalling abstention. Figure~\ref{fig:Example1} (bottom) illustrates this: replacing P\&G with a synthetic organisation name reduces parametric associations, and when back-mapped answers fail to agree across twin worlds, this signals that generation is not reliably grounded in the evidence, and TW enables the model to abstain.

Our contributions are as follows:

\begin{itemize}
\item We reframe abstention in knowledge-intensive reasoning as a test of structural grounding, introducing the \textit{equivariance} criterion: a model whose answers are evidence-grounded should produce outputs that transform consistently under structure-preserving entity substitutions.

\item We propose Twin Worlds, a simple framework that operationalises this criterion by testing whether answers track structural substitutions, not merely whether answers change, using equivariance violations as a training-free inference-time abstention signal.

\item Across four benchmarks and three model backbones, we demonstrate that TW outperforms baselines in abstention reliability and provides diagnostics of answer grounding without requiring evidence-faithfulness annotations.
\end{itemize}

\section{Related Work}

\subsection{Abstention in Large Language Models}

Abstention, where a model explicitly refrains from answering, provides a principled way to handle queries that exceed a model's knowledge or the limits of available evidence~\citep{Kapoor2024LLMMustBeTaught}. Despite its importance, reliable abstention remains challenging: LLMs are primarily trained to maximise benchmark performance rather than recognise their own limitations, and even large models often fail to abstain appropriately~\citep{Kalai2025WhyLLMHallucinate, Kirichenko2025AbstentionBench}. Existing methods approach abstention through fine-tuning~\citep{Zhang2024R-Tuning, StengeleEskin2024Lacie, Cao2024LearnToRefuse, Zhu2025Grait, Huang2025AlleviatingSEAL, Wei2025TruthRL}, uncertainty estimation~\citep{Manakul2023Selfcheckgpt, Yadkori2024ToBelieve, Kim2025Layerwise, Ji2025VerbalUncertainty}, or multi-model agreement~\citep{Feng2024MultilingualAbstain, Feng2024DontHallucinateAbstain, Wen2025Marvel}. More recently, causal approaches analyse the factors influencing abstention decisions~\citep{Sun2025Causalabstain, haoZ0R0LF025, Nguyen2025ABCA}. Despite their differences, these methods largely study abstention in isolation, without accounting for how external evidence influences the generation process. In contrast, TW grounds abstention decisions in the relational structure of the provided evidence, testing whether generation behaviour is consistent with the context itself.

\subsection{Evidence Reliability in Knowledge-Intensive Reasoning}

When models reason over evidence, abstention becomes intertwined with evidence reliability. Models must determine not only whether they know the answer, but whether the available evidence justifies producing one. Risk-based methods estimate when to abstain through context manipulations~\citep{Chen2024ControllingRisk, Wen2024CharacterizingPurtubation}, uncertainty~\citep{Ding2025RoWen}, axiomatic constraints~\citep{Soudani2025AxiomUncertainty}, contrastive decoding~\citep{Kim2025WhenToAbstain}, or agreement~\citep{Goswami2025HaltRAG, hu-etal-2025-removal}. Sufficiency-based approaches evaluate whether retrieved evidence contains the information required to answer~\citep{Sun2025DivideAlign, song2024raghat, Joren2025SufficientContext, Bergeron2025HalluGuard}. However, both strategies test uncertainty or sufficiency without evaluating whether generation is faithful to the provided evidence. A model may remain confident and self-consistent while its answers are driven by parametric activation rather than the provided context. TW addresses this gap directly by testing whether answers are structurally determined by the evidence.

\subsection{Equivariance in Neural Models}

Prior work on equivariance in neural models studies how predictions should transform under structured input transformations, establishing theoretical foundations for group-equivariant architectures~\citep{pmlr-v48-cohenc16, pmlr-v80-kondor18a} and practical methods for complex transformation groups~\citep{pmlr-v139-finzi21a, NEURIPS2020_15231a7c}. In NLP, symmetry failures have been linked to spurious heuristics~\citep{mccoy-etal-2019-right, kassner-schutze-2020-negated}, and behavioural testing frameworks examine model invariance under controlled perturbations~\citep{ribeiro-etal-2020-beyond}. More recent work explores equivariant representations and permutation-based training for language models~\citep{guo-etal-2024-mitigating, li2025large, zhu2025rethinking}. SynthWorlds~\citep{synthworlds} constructs parallel corpora to disentangle reasoning from parametric knowledge, but targets evaluation rather than abstention. These approaches either build equivariance into the model architecture or training procedure, or treat symmetry violations as evidence of poor generalisation. However, whether equivariance can be leveraged to improve LLM reliability at inference time without modifying the model remains unexplored. TW addresses this gap by using equivariance violations as a principled abstention signal in knowledge-intensive reasoning.

\begin{figure*}[t]
    \centering
    \includegraphics[width=1\textwidth]{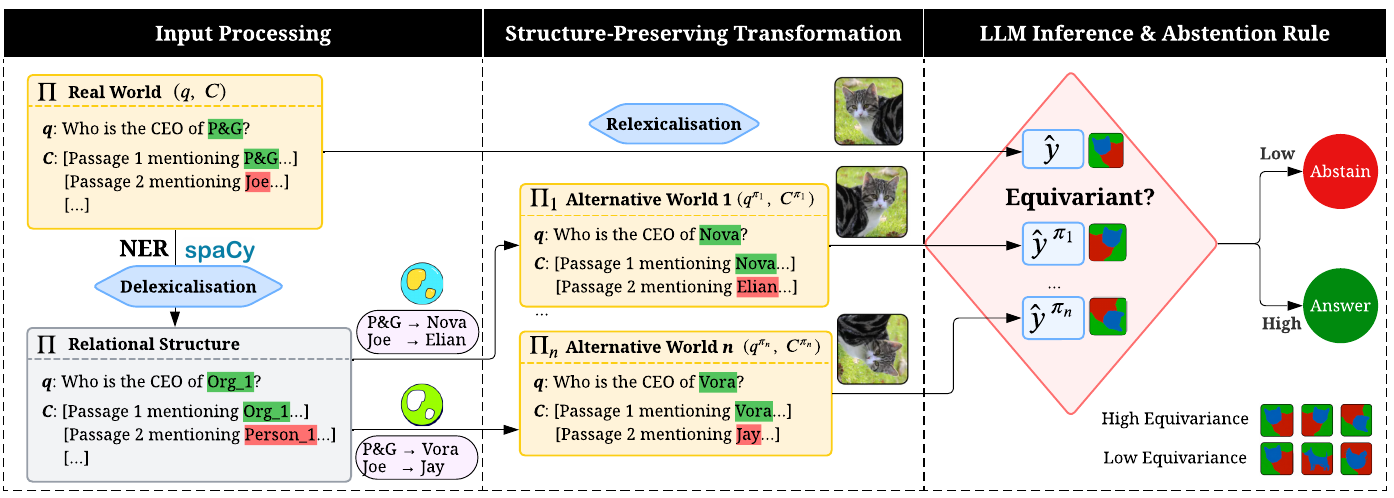}
\caption{TW framework overview. \textit{Left:} Delexicalisation replaces entity mentions with typed placeholders, extracting relational structure. \textit{Middle:} Typed bijective substitutions relexicalise placeholders under $k$ sampled mappings, producing $k$ twin worlds $(q^{\pi_i}, C^{\pi_i})$. \textit{Right:} The equivariance score aggregates whether back-mapped answers $\pi^{-1}(\hat{y}^{\pi_i})$ agree with $\hat{y}$; high equivariance leads to answering, while low equivariance leads to abstention.}
\label{fig:architecture}
\end{figure*}

\section{Methodology}
\label{sec:methodology}
We present \textbf{Twin Worlds (TW)}, a framework for improving the reliability of knowledge-intensive reasoning through equivariance-based abstention.

Let $q$ denote a question, $C = \{c_1, \dots, c_n\}$ a set of evidence passages, and $f$ a language model producing the answer distribution $p_f(y \mid q, C)$. TW evaluates whether answer behaviour is equivariant under structure-preserving transformations of entities in $(q, C)$ and uses violations of this equivariance as a signal for whether the model should answer or abstain. Fig.~\ref{fig:architecture} illustrates this process.

\subsection{Structure-Preserving Transformations}
Let $\mathcal{E}(q, C)$ denote the set of entities appearing in the question and evidence, together with their semantic types such as \textsc{Person}, \textsc{Org}, or \textsc{Date}. A structure-preserving transformation is a mapping defined as follows:
\[
\pi : \mathcal{E}(q, C) \rightarrow \mathcal{E}'(q, C),
\]
such that: (i) type is preserved, so an entity of type $t$ is mapped to another entity of type $t$; (ii) $\pi$ is \textit{bijective}, meaning it is both injective (no two source entities map to the same target) and surjective (every target entity is the image of some source entity), ensuring that the mapping is reversible via $\pi^{-1}$; and (iii) only entity identity changes, while the relational structure remains unchanged.

Applying $\pi$ to $(q, C)$ yields a transformed pair $(q^\pi, C^\pi)$ in which all entity mentions are replaced consistently. We realise these transformations in two steps. First, we \textit{delexicalise} the question and evidence by replacing entity mentions with typed placeholders, producing a canonical representation that preserves relational structure while removing lexical identity. Second, we \textit{relexicalise} under a sampled transformation, replacing each placeholder with a type-matched synthetic entity through a one-to-one reversible mapping. The resulting transformed inputs form \textit{twin worlds}: they preserve the relational structure of the evidence while reducing lexical cues that may activate parametric priors.

\subsection{Equivariance as a Grounding Criterion}
We formalise evidence-grounded reasoning through \textbf{equivariance}. A model is equivariant under transformation $\pi$ if applying $\pi$ to the input induces a corresponding transformation in the answer: the answer in the transformed input, mapped back through $\pi^{-1}$, should agree with the answer in the original input. Intuitively, returning to the example in Figure~\ref{fig:Example1}, if the evidence supports the answer, replacing an entity in the evidence with a synthetic entity such as \textsc{Elian Voss} should produce a correspondingly transformed answer; if the back-mapped answer no longer agrees with the original, this indicates that the answer is not reliably determined by the relational structure of the evidence.

Let $\hat{y} = \arg\max_y p_f(y \mid q, C)$ and $\hat{y}^{\,\pi} = \arg\max_y p_f(y \mid q^\pi, C^\pi)$. When the answer contains an entity participating in $\pi$, $\pi^{-1}$ maps it back through the substitution used to transform the input. For answers that do not participate in the substitution mapping, such as scalar values or relational predicates, $\pi^{-1}$ acts as the identity. We define a per-transformation equivariance score as follows:
\[
S_\pi(q, C) = \mathrm{Agree}\!\left(\pi^{-1}(\hat{y}^{\,\pi}),\, \hat{y}\right),
\]
where $\mathrm{Agree}(\cdot,\cdot)$ is a task-level agreement function. In the simplest case, this reduces to exact match:
\[
S_\pi(q, C) = \mathbf{1}\!\left[\pi^{-1}(\hat{y}^{\,\pi}) = \hat{y}\right].
\]

The back-mapping $\pi^{-1}$ distinguishes equivariance from methods based on an \textit{invariance} criterion. Invariance-based methods produce outputs that remain unchanged under input transformations: confidence- and consistency-based methods test whether $\hat{y}^{\,\pi} = \hat{y}$~\citep{wang2023selfconsistency}; perturbation methods test whether $\hat{y}^{\,\pi} \approx \hat{y}$~\citep{gao-etal-2024-spuq}; counterfactual methods test whether $\hat{y}$ changes when context is removed or replaced~\citep{Chen2024ControllingRisk}. Equivariance instead requires outputs to transform correspondingly: $\pi^{-1}(\hat{y}^{\,\pi}) = \hat{y}$~\citep{geva-etal-2023-dissecting}. A model relying on parametric memory may pass invariance-based tests yet fail equivariance because its answers do not track the substitution mapping.

TW constructs a set of $k$ twin worlds $W = \{\pi_1, \dots, \pi_k\}$ by sampling structure-preserving transformations over $\mathcal{E}(q, C)$. Each twin world $\pi_i$ produces a transformed input $(q^{\pi_i}, C^{\pi_i})$ by applying the substitution consistently across all entity mentions in the question and evidence passages. Twin worlds reduce parametric priors by using procedurally generated synthetic values as replacement entities, such as \textsc{Nova} for a location or \textsc{Elian Voss} for a person. This design encourages equivariant behaviour across twin worlds to reflect relational reasoning from the evidence rather than incidental parametric associations.

\subsection{Equivariance-Based Abstention}
TW computes an equivariance score by averaging over the set of twin worlds:
\[
s(q, C) = \frac{1}{|W|} \sum_{\pi \in W} S_\pi(q, C).
\]
This score measures whether the answer is reliably determined by the evidence. TW applies the following abstention rule:
\[
\hat{y}_{\mathrm{out}} =
\begin{cases}
\hat{y}, & \text{if } s(q, C) \ge \tau \\[4pt]
\texttt{abstain}, & \text{otherwise,}
\end{cases}
\]
where $\tau$ is a threshold tuned on a held-out validation set. When the equivariance score falls below the threshold, the answer is not reliably determined by the relational structure of the provided evidence, and TW abstains. TW does not attempt to distinguish whether such a failure arises from the model, irrelevant or erroneous context, or another source; rather, it signals that the answer cannot be reliably justified from the given evidence.

\section{Experiments \& Results}
\begin{table*}[t]
\centering
\scriptsize
\setlength{\tabcolsep}{4.2pt} 
\renewcommand{\arraystretch}{1.1}
\input{main-results}
\caption{Results of TW and baselines across benchmark datasets. Each section compares methods with the same backbone. \textbf{Bold} indicates the highest score and \underline{underline} the second highest. A hyphen (-) denotes zero F1 on FaithEval where all questions are unanswerable; performance there is measured by RS and Acc. Higher AR indicates more frequent abstention and may reflect over-conservatism rather than genuine grounding.}
\label{tab:experiment-results}
\end{table*}

\subsection{Experimental Settings}
\paragraph{Datasets}
We evaluate TW on four benchmarks. HotpotQA~\citep{Yang2018HotpotQA} tests multi-hop reasoning with distractor passages. MIRAGE~\citep{Park2025Mirage} aggregates open-domain QA datasets, including PopQA~\citep{mallen-etal-2023-trust} and TriviaQA~\citep{Joshi2017TriviaQA}, enabling analysis of parametric versus contextual knowledge use. FaithEval~\citep{Ming2025FaithEval} introduces unanswerable and misleading contexts to assess robustness against unsupported answering. FEVER~\citep{Thorne2018Fever} evaluates fact verification, where models should abstain on \textit{Not Enough Info} instances when evidence is insufficient. Rather than prompting the model to classify claims into predefined labels, we treat FEVER as a generative task: the model is prompted to answer each claim as a question given the retrieved evidence, and TW evaluates equivariance via $\pi^{-1}$ exactly as in the QA setting. Dataset sampling and construction details are provided in Appendix~\ref{appx:datasets}.

\paragraph{Baselines} 
We compare TW against seven baselines. Zero-shot~\citep{Kojima2022Zeroshot} measures inherent abstention behaviour under direct prompting. Risk-based methods estimate when to abstain from observable signals: AbstentionBench~\citep{Kirichenko2025AbstentionBench} uses abstention-oriented prompting, Self-Consistency~\citep{wang2023selfconsistency} detects when agreement across sampled reasoning traces is low, and RC-RAG~\citep{Chen2024ControllingRisk} estimates hallucination risk via counterfactual context manipulations. Sufficiency-based methods assess whether the evidence supports an answer: Context Perturbation~\citep{Wen2024CharacterizingPurtubation} measures answer stability under evidence perturbations, Sufficient Context~\citep{Joren2025SufficientContext} predicts evidence sufficiency via a scoring model, and Contrastive Decoding Abstention~\citep{Kim2025WhenToAbstain} modifies decoding to abstain under weak or unreliable evidence. For methods with publicly available implementations, we use the authors' official repositories and default configurations.

\paragraph{Metrics}
Effective abstention must balance two failure modes: answering when evidence is insufficient and refusing when it is not. Following \citet{Kim2025WhenToAbstain, Madhusudhan2025DoLLMsKnow}, we use a confusion matrix crossing question types (answerable vs.\ unanswerable) with model behaviours (answered correctly, answered incorrectly, or abstained), and report F1, Accuracy (Acc), Reliability Score (RS), and Abstention Rate (AR). RS captures the trade-off between coverage and correctness; AR flags methods that resolve this trade-off by over-refusing rather than improving grounding. Full metric definitions are provided in Appendix~\ref{appx:metrics}.

\paragraph{Twin World Construction}
We first \textit{delexicalise} the question and evidence using spaCy~\citep{Honnibal2023Spacy} to detect entity spans and assign coarse semantic types (\textsc{Person}, \textsc{Org}, \textsc{Location}, \textsc{Date}), replacing each mention with a typed indexed placeholder such as \texttt{Person\_1} or \texttt{Org\_1} and merging repeated mentions through string and alias matching; robustness to NER failures is analysed in App.~\ref{appx:error-analysis}. We then \textit{relexicalise} by instantiating structure-preserving transformations over these placeholders, sampling without replacement from a task-agnostic inventory of synthetic values with no attested referents in standard English; the same inventory is reused across all four benchmarks. Generation and validation details are in App.~\ref{appx:synthetic-entities}. We sample $k = 3$ twin worlds per query. The default threshold $\tau = 0.60$ is tuned on a 500-instance held-out validation set to maximise RS performance.

\paragraph{Model Backbones}
We evaluate TW on GPT-5.1, LLaMA-4-Scout 17B$\times$16E, and Mistral-Small-24B, representing closed-source, MoE, and compact architectures respectively. Detailed experimental setups are provided in Appendix~\ref{appx:experiment-settings}.

\begin{table}[t]
\centering
\scriptsize
\setlength{\tabcolsep}{2.5pt}
\renewcommand{\arraystretch}{1}
\begin{tabular}{p{0.95\columnwidth}}
\toprule
\textbf{Q:} What was found in \colorbox{green!25}{Los Angeles}? \\
\textbf{E:} [...] \colorbox{blue!20}{Trevor Valle} discusses the teeth and lower jaw of a mammoth fossil. Now, at least 10,000 years later, visitors in \colorbox{green!25}{Los Angeles} can see the remains of a mammoth [...] \\
\midrule
\textbf{Twin World 1:} {Los Angeles} $\mapsto$ {Nova},\; {Trevor Valle} $\mapsto$ {Elian Voss} \\
\textbf{Output:} no information \hfill $S_{\pi_1} = 0$ \\
\midrule
\textbf{Twin World 2:} {Los Angeles} $\mapsto$ {Valdecor},\; {Trevor Valle} $\mapsto$ {Orin Vael} \\
\textbf{Output:} no information \hfill $S_{\pi_2} = 0$ \\
\midrule
\textbf{Twin World 3:} {Los Angeles} $\mapsto$ {Mirevan},\; {Trevor Valle} $\mapsto$ {Casen Drel} \\
\textbf{Output:} no answer \hfill $S_{\pi_3} = 0$ \\
\midrule
\textbf{Final Answer:} $s(q,C) = 0.00 < \tau = 0.60 \Rightarrow \textbf{NIL (Abstain)}$ \\
\bottomrule
\end{tabular}
\caption{TW on a representative unanswerable case. Across three twin worlds, all twin answers fail to agree with the original answer ``mammoth fossil'', yielding $s = 0.00$ and triggering abstention.}
\label{tab:tw-abstention-example}
\end{table}

\subsection{Main Results}
Table~\ref{tab:experiment-results} reports the performance of TW against all baselines. TW achieves the strongest overall performance on HotpotQA and MIRAGE across all three backbones, attaining the highest F1 and Accuracy in all six dataset--backbone combinations and the highest RS in four of six cases, while maintaining near-zero abstention rates (AR $\leq$ 0.029). In contrast, RC-RAG reaches AR values between 0.27 and 0.54, while Context Perturbation reaches up to 0.25, and both underperform TW on F1 and RS. These results indicate that equivariance-based abstention improves answer quality without over-refusing on answerable questions.

On FaithEval, where all questions are unanswerable, TW achieves the best Accuracy and AR on LLaMA-4 and Mistral-Small and the second-best results on GPT-5.1. Sufficient Context abstains more aggressively on GPT-5.1 but also substantially over-abstains on answerable benchmarks, reaching AR 0.496 on MIRAGE. This illustrates the difficulty of balancing abstention on unanswerable instances against coverage on answerable ones. TW instead bases its abstention signal on whether answers track the relational structure of the evidence across structure-preserving transformations. On FEVER, TW achieves the best F1 across all three backbones, suggesting that equivariance-based abstention generalises beyond question answering to fact verification.

Table~\ref{tab:tw-abstention-example} illustrates a representative case. Under structure-preserving substitutions, the answer fails to track the substitution mapping, and TW correctly abstains, indicating that the original answer is not reliably determined by the relational structure of the provided evidence.

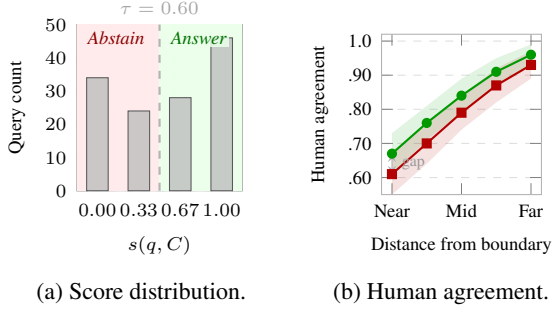
\begin{figure}[t]
  \centering

  \begin{subfigure}{0.49\linewidth}
  \begin{tikzpicture}
  \begin{axis}[
      width=\linewidth,
      height=\linewidth,
      ymin=0, ymax=50,
      xmin=-0.5, xmax=3.5,
      xlabel={$s(q,C)$},
      ylabel={Query count},
      xlabel style={font=\scriptsize},
      ylabel style={font=\scriptsize},
      tick label style={font=\scriptsize},
      xtick={0,1,2,3},
      xticklabels={$0.00$,$0.33$,$0.67$,$1.00$},
      x tick label style={
          font=\scriptsize,
          yshift=-1pt
      },
      ytick={0,10,20,30,40,50},
      ymajorgrids=true,
      xmajorgrids=false,
      grid style={dashed, gray!25},
      axis line style={gray!60},
      clip=false
  ]

  \addplot[fill=red!8, draw=none, forget plot]
      coordinates {(-0.5,0) (1.5,0) (1.5,50) (-0.5,50) (-0.5,0)};

  \addplot[fill=green!8, draw=none, forget plot]
      coordinates {(1.5,0) (3.5,0) (3.5,50) (1.5,50) (1.5,0)};

  \addplot[dashed, thick, gray!60, forget plot]
      coordinates {(1.5,0) (1.5,50)};

  \node[font=\scriptsize, gray!70, anchor=south]
      at (axis cs:1.5,50) {$\tau=0.60$};

  \addplot[
      ybar,
      bar width=0.52,
      fill=gray!50,
      draw=gray!70!black,
      fill opacity=0.80
  ] coordinates {
      (0,34)
      (1,24)
      (2,28)
      (3,46)
  };

  \node[
      font=\scriptsize\itshape,
      red!55!black,
      fill=red!8,
      fill opacity=0.85,
      text opacity=1,
      inner sep=1pt
  ] at (axis cs:0.45,46) {Abstain};

  \node[
      font=\scriptsize\itshape,
      green!45!black,
      fill=green!8,
      fill opacity=0.85,
      text opacity=1,
      inner sep=1pt
  ] at (axis cs:2.45,46) {Answer};

  \end{axis}
  \end{tikzpicture}
  \caption{Score distribution.}
  \label{fig:score-distribution}
  \end{subfigure}
  \hfill
  \begin{subfigure}{0.48\linewidth}
  \begin{tikzpicture}
  \begin{axis}[
      width=\linewidth,
      height=\linewidth,
      xmin=-0.3, xmax=4.3,
      ymin=0.55, ymax=1.02,
      xlabel={Distance from boundary},
      ylabel={Human agreement},
      xlabel style={font=\scriptsize},
      ylabel style={font=\scriptsize},
      tick label style={font=\scriptsize},
      xtick={0, 2, 4},
      xticklabels={\scriptsize Near, \scriptsize Mid, \scriptsize Far},
      ytick={0.6, 0.7, 0.8, 0.9, 1.0},
      yticklabels={.60, .70, .80, .90, 1.0},
      ymajorgrids=true,
      grid style={dashed, gray!25},
      axis line style={gray!60},
      legend style={
          at={(0.50, -0.14)},
          anchor=north,
          draw=gray!40,
          fill=white,
          font=\tiny,
          legend columns=2,
          legend cell align=left,
          column sep=4pt,
          inner sep=2pt,
      },
      clip=false
  ]

  \addplot[name path=ans_upper, draw=none, forget plot] coordinates {
      (0, 0.73) (1, 0.81) (2, 0.89) (3, 0.95) (4, 0.99)
  };

  \addplot[name path=ans_lower, draw=none, forget plot] coordinates {
      (0, 0.61) (1, 0.71) (2, 0.79) (3, 0.87) (4, 0.93)
  };

  \addplot[
      fill=green!60!black,
      fill opacity=0.12,
      draw=none,
      forget plot
  ] fill between[of=ans_upper and ans_lower];

  \addplot[
      thick,
      mark=*,
      mark size=1.5pt,
      color=green!60!black
  ] coordinates {
      (0, 0.67) (1, 0.76) (2, 0.84) (3, 0.91) (4, 0.96)
  };

  \addplot[name path=abs_upper, draw=none, forget plot] coordinates {
      (0, 0.67) (1, 0.76) (2, 0.84) (3, 0.92) (4, 0.97)
  };

  \addplot[name path=abs_lower, draw=none, forget plot] coordinates {
      (0, 0.55) (1, 0.64) (2, 0.74) (3, 0.82) (4, 0.89)
  };

  \addplot[
      fill=red!70!black,
      fill opacity=0.12,
      draw=none,
      forget plot
  ] fill between[of=abs_upper and abs_lower];

  \addplot[
      thick,
      mark=square*,
      mark size=1.5pt,
      color=red!70!black
  ] coordinates {
      (0, 0.61) (1, 0.70) (2, 0.79) (3, 0.87) (4, 0.93)
  };

  \draw[gray!60, <->, >=stealth, thin]
      (axis cs:0, 0.61) -- (axis cs:0, 0.67)
      node[midway, right, font=\tiny, gray!80] {gap};

  \end{axis}
  \end{tikzpicture}
  \caption{Human agreement.}
  \label{fig:human-agreement}
  \end{subfigure}

  \caption{(a) Distribution of equivariance scores on GPT-5.1. With $k=3$, scores take the discrete values $\{0, 0.33, 0.67, 1.00\}$; most queries lie away from the decision boundary at $\tau=0.60$. (b) Human--TW agreement increases monotonically with distance from the decision boundary, indicating stronger alignment on cases with more decisive equivariance scores.}
  \label{fig:combined-human-eval}
\end{figure}

\subsection{Human Evaluation}
We conduct human evaluation to validate two aspects of TW: the reliability of structure-preserving transformations in preserving relational structure and the alignment of abstention decisions with human judgement.

\paragraph{Setup} We recruit native English speakers via Amazon Mechanical Turk with $\geq$98\% approval rate and $\geq$500 completed HITs. Each instance is evaluated by three independent annotators with attention checks. We sample 200 instances from MIRAGE and HotpotQA, balanced across entity types. Detailed annotation instructions and scoring rubrics are provided in Appendix~\ref{appx:human-eval}.

\paragraph{Structure Preservation} Annotators rate how well transformed inputs preserve relational structure on a 5-point Likert scale. Results indicate a mean rating of 4.61 (SD = 0.48), with 99.1\% of ratings $\geq 4$, suggesting that structure-preserving transformations reliably preserve relational structure while altering entity identity.

\paragraph{Abstention Alignment} Annotators judge whether the evidence contains sufficient information to answer the question, without access to TW outputs. TW agrees with human judgement on 86\% of answerable and 81\% of unanswerable instances, yielding 83.5\% overall agreement (Fleiss' $\kappa = 0.78$). Fig.~\ref{fig:combined-human-eval} shows that alignment increases monotonically with distance from $\tau$, exceeding 90\% for queries far from the boundary. Residual disagreement concentrates near the boundary, where evidential support is more ambiguous, suggesting that equivariance scores align more strongly with human judgement away from the decision boundary.

\subsection{Probing Structural Grounding}

We validate TW from two complementary angles: behavioural and representational, using 500 instances per condition from MIRAGE and HotpotQA with passage faithfulness annotations.

\paragraph{Behavioural Analysis.} We evaluate the World-Tracking Rate (WTR), a metric measuring how consistently back-mapped answers agree with the original answer across twin worlds, under two conditions: \textit{faithful} passages that uniquely support the answer, and \textit{unfaithful} passages that mention relevant entities without supporting it. We compute:
\[
\mathrm{WTR}(q, C) = \frac{1}{|W|} \sum_{\pi \in W} \mathbf{1}\!\left[\pi^{-1}(\hat{y}^{\pi}) = \hat{y}\right].
\]
We compare typed bijective transformations with four alternatives: \textit{untyped substitution}, which permits cross-type replacements; \textit{counterfactual substitution}, which uses real-world alternatives of the same type; \textit{surface paraphrase}, which rewrites entity mentions without substituting their identity; and \textit{random replacement}, which replaces entity spans with random tokens. Table~\ref{tab:wtr} shows that typed bijective transformations produce the clearest separation between faithful and unfaithful evidence across all backbones. Counterfactual substitution yields substantially higher WTR under unfaithful evidence, consistent with real-world replacements retaining parametric associations; surface paraphrase produces similar WTR regardless of faithfulness; and random replacement degrades WTR under both conditions.

\begin{figure}[t]
\centering
\begin{tikzpicture}
\begin{axis}[
    width=\columnwidth,
    height=0.4\columnwidth,
    xlabel={Transformer layer},
    ylabel={Token probability},
    xlabel style={font=\small},
    ylabel style={font=\small},
    tick label style={font=\scriptsize},
    xmin=1, xmax=32,
    ymin=0, ymax=0.35,
    xtick={1, 8, 16, 24, 32},
    ytick={0, 0.10, 0.20, 0.30},
    ymajorgrids=true,
    grid style={dashed, gray!25},
    axis line style={gray!60},
    legend style={
        at={(0.5, -0.22)},
        anchor=north,
        draw=gray!40,
        fill=white,
        font=\tiny,
        legend columns=2,
        column sep=6pt,
        inner sep=2pt,
    },
]

\addplot[thick, color=red!70!black, mark=none] coordinates {
    (1,0.08) (4,0.10) (8,0.14) (12,0.19) (16,0.26) (20,0.30) (24,0.28) (28,0.24) (32,0.21)
};
\addlegendentry{Real + unfaithful}

\addplot[thick, color=orange!80!black, mark=none, dashed] coordinates {
    (1,0.07) (4,0.09) (8,0.12) (12,0.16) (16,0.20) (20,0.22) (24,0.20) (28,0.17) (32,0.15)
};
\addlegendentry{Real + faithful}

\addplot[thick, color=blue!60!black, mark=none] coordinates {
    (1,0.03) (4,0.04) (8,0.05) (12,0.06) (16,0.07) (20,0.07) (24,0.06) (28,0.05) (32,0.05)
};
\addlegendentry{Synthetic + unfaithful}

\addplot[thick, color=green!50!black, mark=none, dashed] coordinates {
    (1,0.03) (4,0.04) (8,0.04) (12,0.05) (16,0.06) (20,0.06) (24,0.05) (28,0.05) (32,0.04)
};
\addlegendentry{Synthetic + faithful}

\end{axis}
\end{tikzpicture}
\caption{Probability of entity-associated tokens across Mistral-Small-24B layers. Synthetic substitutions consistently reduce activation relative to the corresponding real-entity inputs, with the largest activation observed for real entities under unfaithful evidence.}
\label{fig:activation}
\end{figure}
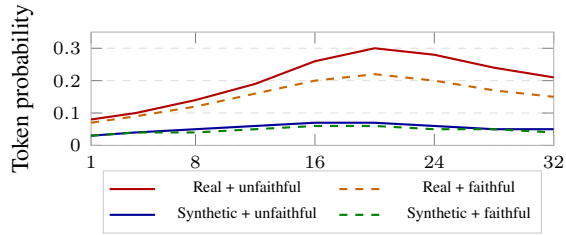

\paragraph{Activation Analysis.} We examine whether synthetic substitutions reduce entity-associated parametric activation rather than merely altering surface form. For each instance, we prompt Mistral-Small-24B with the question and no evidence and define \textit{entity-associated tokens} as the top-10 tokens in the resulting closed-book output distribution for the original entity input. We then measure the probability of these same tokens across transformer layers under original and synthetic entity inputs using the logit lens. Figure~\ref{fig:activation} shows that synthetic substitutions consistently reduce entity-associated activation across layers, while original entities with unfaithful evidence exhibit higher activation, particularly in later layers. Together with the behavioural results, this provides evidence that synthetic substitutions attenuate lexical triggers associated with parametric memory and supports the interpretation of equivariance violations as a signal that an answer is not reliably determined by the relational structure of the provided evidence.

\begin{table}[t]
\centering
\scriptsize
\setlength{\tabcolsep}{2.8pt}
\renewcommand{\arraystretch}{1}
\begin{tabular}{l|cc|cc|cc}
\toprule
& \multicolumn{2}{c|}{\textbf{GPT-5.1}} & \multicolumn{2}{c|}{\textbf{LLaMA-4}} & \multicolumn{2}{c}{\textbf{Mistral-Small}} \\
\textbf{Strategy} & \textbf{Faith.} & \textbf{Unfaith.} & \textbf{Faith.} & \textbf{Unfaith.} & \textbf{Faith.} & \textbf{Unfaith.} \\
\midrule
\textbf{Typed Bijection} & \textbf{.804} & .215 & \textbf{.780} & \textbf{.230} & \textbf{.761} & \textbf{.243} \\
\midrule
Untyped Sub.      & .721 & .401 & .701 & .385 & .685 & .390 \\
Counterfactual    & .687 & .475 & .670 & .460 & .655 & .452 \\
Paraphrase        & .811 & .800 & .799 & .790 & .791 & .780 \\
Random Repl.      & .340 & .\textbf{210} & .325 & .305 & .310 & .301 \\
\bottomrule
\end{tabular}%
\caption{WTR across evidence conditions, substitution strategies, and model backbones. Higher WTR is desirable under faithful evidence and lower WTR under unfaithful evidence; the desired behaviour is therefore a large faithful--unfaithful separation.}
\label{tab:wtr}
\end{table}

\subsection{Ablation Studies}
We ablate five design choices on a balanced sample of 500 instances (250 answerable, 250 unanswerable) drawn from all four datasets. Results appear in Table~\ref{tab:ablation}.

\paragraph{Number of Twin Worlds} We sweep $k \in \{1, 2, 3, 4, 6\}$ and find that $k=1$ reduces RS by 0.044 across all three backbones, as a single substitution yields a less stable equivariance estimate. Performance improves up to $k=3$. Increasing to $k=6$ yields consistent but negligible gains on most metrics compared with the doubled inference cost, while GPT-5.1 RS decreases slightly. We therefore use $k=3$ as the default.

\paragraph{Substitution Properties} Removing type constraints allows cross-type substitutions, introducing type-mismatched inputs that confound the abstention signal and reducing RS by 0.048--0.051 across backbones. Removing bijectivity breaks the one-to-one mapping between source and target entities, collapsing structural distinctions between entity roles and weakening diagnostic separation. Replacing synthetic values with high-frequency Wikipedia entities produces the largest RS drop (0.103--0.115) across backbones, suggesting that non-referential synthetic entity construction is an important design choice.

\paragraph{Abstention Mechanism} Removing back-mapping reduces RS by 0.067--0.076 across backbones, demonstrating the importance of testing equivariance via $\pi^{-1}$. Without back-mapping, the framework reduces to a perturbation stability check without an explicit structural correspondence between transformed and original answers.

\begin{table}[t]
\centering
\scriptsize
\setlength{\tabcolsep}{2.6pt}
\renewcommand{\arraystretch}{1}
\begin{tabular}{l|ccc|ccc|ccc}
\toprule
& \multicolumn{3}{c|}{\textbf{GPT-5.1}} 
& \multicolumn{3}{c|}{\textbf{LLaMA-4}} 
& \multicolumn{3}{c}{\textbf{Mistral-Small}} \\
\textbf{Configuration} 
& \textbf{F1} & \textbf{Acc} & \textbf{RS} 
& \textbf{F1} & \textbf{Acc} & \textbf{RS} 
& \textbf{F1} & \textbf{Acc} & \textbf{RS} \\
\midrule
TW (Full, $k=3$) 
& .921 & .903 & \textbf{.971} 
& .891 & .873 & .956 
& .867 & .849 & .934 \\
\midrule
$k=1$ 
& .884 & .859 & .927 
& .853 & .831 & .912 
& .829 & .808 & .890 \\

$k=6$ 
& \textbf{.922} & \textbf{.904} & .967 
& \textbf{.892} & \textbf{.874} & \textbf{.957} 
& \textbf{.868} & \textbf{.850} & \textbf{.935} \\
\midrule
No type constraint 
& .874 & .843 & .921 
& .843 & .819 & .908 
& .819 & .791 & .883 \\

No bijectivity 
& .881 & .851 & .928 
& .851 & .826 & .913 
& .828 & .801 & .891 \\

Wiki entities 
& .821 & .793 & .868 
& .791 & .764 & .841 
& .768 & .742 & .831 \\
\midrule
No back-mapping 
& .849 & .819 & .904 
& .819 & .794 & .881 
& .796 & .771 & .858 \\
\bottomrule
\end{tabular}
\caption{Ablation results across all three backbones, averaged across all four datasets. \textbf{Bold} indicates the highest score in each column.}
\label{tab:ablation}
\end{table}

\subsection{Analysis of Hard Abstention Cases}
A key claim of TW is that equivariance captures failure cases that may not be identified by uncertainty or sufficiency signals. We test this by constructing a hard subset of 200 unanswerable instances drawn from all four datasets, selected such that Self-Consistency~\citep{wang2023selfconsistency} agreement is high ($\geq 0.8$) and Sufficient Context~\citep{Joren2025SufficientContext} score is high ($\geq 0.5$). These are cases in which both signals favour answering despite the absence of sufficient evidence. Table~\ref{tab:hard-cases} shows that Self-Consistency achieves AR of 0.09 and Sufficient Context achieves AR of 0.11, whereas TW achieves AR of 0.71. This indicates that equivariance provides a complementary abstention signal in cases where confidence and apparent evidence sufficiency are both high.

Table~\ref{tab:hard-case-example} illustrates a representative case. The passage states that Gareth Bale spent much of his career as a winger but does not establish his position specifically in his final season. Self-Consistency produces identical answers across five traces and Sufficient Context assigns a score of 0.51, leading both methods to answer. Across three twin worlds, TW produces inconsistent back-mapped answers ($s = 0.33 < \tau$) and abstains, indicating that the original answer is not reliably determined by the provided evidence.

\begin{table}[t]
\centering
\scriptsize
\setlength{\tabcolsep}{8pt}
\renewcommand{\arraystretch}{1}
\begin{tabular}{l|c}
\toprule
\textbf{Method} & \textbf{AR} \\
\midrule
AbstentionBench & 0.18 \\
Self-Consistency & 0.09 \\
RC-RAG & 0.31 \\
Context Perturbation & 0.24 \\
Sufficient Context & 0.11 \\
Contrastive Decoding & 0.22 \\
\textbf{TW (Ours)} & \textbf{0.71} \\
\bottomrule
\end{tabular}
\caption{Abstention rate on 200 unanswerable instances drawn from all four datasets, where both Self-Consistency agreement and Sufficient Context score are high.}
\label{tab:hard-cases}
\end{table}

\begin{table}[t]
\centering
\scriptsize
\setlength{\tabcolsep}{3pt}
\renewcommand{\arraystretch}{1.05}
\begin{tabular}{p{0.36\columnwidth} p{0.38\columnwidth} r}
\toprule
\multicolumn{3}{l}{\textbf{Q:} What position did \colorbox{green!25}{Gareth Bale} play in his final season?} \\
\multicolumn{3}{p{0.95\columnwidth}}{\textbf{E:} \colorbox{green!25}{Gareth Bale} announced his retirement from professional football in January 2023, having spent much of his career as a \textit{winger}, citing his desire to spend more time with his family.} \\
\multicolumn{3}{l}{\textbf{Gold:} unanswerable} \\
\midrule
\multicolumn{3}{l}{\textbf{Self-Consistency:} 5/5 traces answer ``winger'' \hfill agreement $= 1.0$} \\
\multicolumn{3}{l}{\textbf{Sufficient Context:} passage score $= 0.51$ \hfill both baselines answer} \\
\midrule
\textbf{TW 1:} {Bale} $\mapsto$ {Elian Voss}
& \textbf{Output:} winger
& $S_{\pi_1} = 1$ \\

\textbf{TW 2:} {Bale} $\mapsto$ {Orin Vael}
& \textbf{Output:} no information
& $S_{\pi_2} = 0$ \\

\textbf{TW 3:} {Bale} $\mapsto$ {Casen Drel}
& \textbf{Output:} no information
& $S_{\pi_3} = 0$ \\
\midrule
\multicolumn{3}{l}{$s(q,C) = 0.33 < \tau = 0.60 \Rightarrow \textbf{TW abstains}$} \\
\bottomrule
\end{tabular}
\caption{A hard unanswerable case where Self-Consistency and Sufficient Context both favour answering. TW detects inconsistent back-mapped answers and abstains because the answer is not reliably determined by the provided evidence.}
\label{tab:hard-case-example}
\end{table}

\subsection{Additional Analyses}
Due to space constraints, we provide additional analyses in the appendices. Parameter sensitivity analysis (Appendix~\ref{appx:parameter-analysis}) shows that performance improves up to $k=3$ and changes only marginally with additional twin worlds, supporting $k=3$ as the default. Error analysis (Appendix~\ref{appx:error-analysis}) shows that TW remains robust to moderate NER failures, while most ambiguity is concentrated near the equivariance decision boundary. Computational cost analysis (Appendix~\ref{app:compute-cost}) shows that TW achieves higher reliability than comparable multi-pass baselines at similar latency. Case studies (Appendix~\ref{appx:case-studies}) illustrate that TW preserves answers when they remain structurally supported by the evidence and abstains when this support breaks under entity substitution.

\section{Conclusion}

We introduce Twin Worlds (TW), a framework that probes evidence-grounded reasoning through equivariance. If answers are determined by the evidence, they transform consistently under structure-preserving entity substitutions, and violations of this criterion signal abstention. Rather than asking whether a model is uncertain or whether evidence is sufficient, TW asks whether an answer is structurally determined by the provided evidence via structure-preserving transformations that reduce parametric priors. Across four benchmarks and three backbones, TW outperforms uncertainty- and sufficiency-based baselines while maintaining strong answer quality on answerable instances. Future work will extend TW beyond named entities to other structured elements, while also exploring multilingual settings and longer-form generation.

\section*{Limitations}

TW is most effective for entity-grounded reasoning tasks where questions, answers, and evidence are structured around named entities. Tasks with limited entity structure, such as mathematical or procedural reasoning, are outside the current scope, and TW may also be less straightforward in domains where entity recognition is unreliable. Extending the framework to finer-grained structured elements, such as numbers, spans, clauses, or domain-specific terms, may broaden its applicability to longer-form and more open-ended generation. Current evaluation focuses on English-language benchmarks; multilingual settings may require language-specific substitution strategies that account for morphological variation, transliteration, and language-dependent entity representations.

\section*{Ethical considerations}

This work aims to improve abstention behaviour in knowledge-intensive reasoning to reduce hallucinations and unsupported responses. All benchmarks used, including HotpotQA, MIRAGE, FaithEval, and FEVER, are publicly available and contain no personal or sensitive information. Human evaluation is conducted via Amazon Mechanical Turk, where annotators assess evidential sufficiency and structure preservation in factual question answering tasks. The proposed framework encourages abstention when evidential support is weak, mitigating risks associated with misinformation rather than introducing new ethical concerns.

\section*{Acknowledgments}

This research is supported in part by the Australian Research Council Discovery Project DP260104188 and the RMIT University School of Computing Technologies research support package.

\bibliography{custom}

\newpage
\appendix

\section{Appendix: Experimental Settings}
\label{appx:experiment-settings}

\subsection{Dataset Details}
\label{appx:datasets}

\paragraph{HotpotQA.}
We use the distractor setting from the full Wikipedia split~\citep{Yang2018HotpotQA}, which provides ten passages per query, of which two are gold supporting passages and eight are distractors. We sample 2,000 instances for evaluation, stratified to balance bridge and comparison question types.

\paragraph{MIRAGE.}
We use 2,000 instances from the evaluation split of \citet{Park2025Mirage}, which provides passage-level faithfulness annotations distinguishing passages that uniquely support the answer from those that mention relevant entities without supporting it.

\paragraph{FaithEval.}
We use 2,000 instances from the unanswerable split~\citep{Ming2025FaithEval}, in which all instances lack sufficient evidence for answering. Because no instance has a valid answer, F1 is zero by construction and is reported as a dash in Table~\ref{tab:experiment-results}. Performance is therefore measured primarily by AR and Acc, which coincide on this all-unanswerable subset.

\paragraph{FEVER.}
We use 2,000 instances from the \textit{Supports}, \textit{Refutes}, and \textit{Not Enough Info} labels of \citet{Thorne2018Fever}, sampled to balance the three label classes. \textit{Not Enough Info} instances are treated as unanswerable, since the claim cannot be verified from the retrieved evidence alone. \textit{Supports} and \textit{Refutes} instances are treated as answerable, with the model expected to produce the corresponding verdict.

Since FEVER does not provide retrieved evidence, we use BM25~\citep{Robertson2009BM25} to retrieve the top three documents from a Wikipedia dump dated 01 November 2025. This retrieval setting is used identically for TW and all baselines to ensure fair comparison.

Applying TW to FEVER requires no adaptation to the core equivariance criterion. Rather than prompting the model to classify claims directly into predefined labels, we treat FEVER as a generative task: the model is prompted to answer each claim as a question given the retrieved evidence. Structure-preserving transformations are applied to entity mentions in both the claim and evidence passages. For yes/no outputs that do not participate in the substitution mapping, $\pi^{-1}$ acts as the identity. Equivariance therefore requires the semantic verdict to remain consistent between the original input and its twin worlds. When $s(q,C)<\tau$, TW abstains; abstention is mapped to \textit{Not Enough Info} at evaluation time.

\subsection{Validation Set and Threshold Selection}
\label{appx:threshold}

The threshold $\tau=0.60$ is selected on a held-out validation set of 500 instances drawn from all four datasets, stratified to balance answerable and unanswerable instances within each dataset. All validation instances are distinct from the test instances used in Table~\ref{tab:experiment-results}; no instance appears in both splits.

With the default $k=3$, the equivariance score takes values in
$\{0,\frac{1}{3},\frac{2}{3},1\}$. Thus, $\tau=0.60$ requires agreement in at least two of the three twin worlds.

\subsection{Synthetic Entity Inventory}
\label{appx:synthetic-entities}

\paragraph{Generation procedure.}
The synthetic entity inventory is organised around the four coarse types derived from spaCy's NER pipeline (\textsc{Person}, \textsc{Org}, \textsc{Location}, \textsc{Date}), with a separate pool of values for each type. The same task-agnostic inventory is reused across all four benchmarks.

Values within each pool are procedurally generated to satisfy three properties: (i) no attested referent in standard English, verified by checking against a Wikipedia title index and a frequency-filtered web corpus; (ii) phonological plausibility within the entity type, so that \textsc{Person} values follow first-name/last-name structure and \textsc{Location} values follow place-name phonotactics; and (iii) no surface overlap with any entity appearing in the evaluation benchmarks, verified by exact string matching against all entity spans extracted by spaCy across the full evaluation set.

Concretely, \textsc{Person} values are generated by randomly combining syllable sequences drawn from a phoneme inventory constrained to produce name-like strings (e.g., \textsc{Elian Voss}, \textsc{Orin Vael}, \textsc{Casen Drel}). \textsc{Location} values follow a similar procedure constrained to produce place-name-like strings (e.g., \textsc{Nova}, \textsc{Valdecor}, \textsc{Mirevan}). \textsc{Org} values are generated as two-part strings combining a synthetic adjective and a generic organisational noun (e.g., \textsc{Verathen Group}). \textsc{Date} values are not replaced with synthetic strings; instead, dates are shifted by a fixed offset sampled uniformly from $\pm[1,10]$ years, preserving temporal structure while displacing the specific value.

We relexicalise typed placeholders into synthetic entities rather than directly querying the model with placeholders such as \texttt{Person\_1}. This preserves linguistically plausible surface forms while avoiding unnatural token patterns that could affect generation independently of grounding.

\paragraph{Validation.}
To confirm that synthetic values do not carry residual parametric associations, we query each backbone model with prompts of the form \textit{``What do you know about [synthetic value]?''} and \textit{``Is [synthetic value] a real [person/place/organisation]?''} A value is accepted if all three backbones respond with an explicit statement of non-recognition. Values that elicit factual claims or confident associations are discarded and regenerated.

\subsection{Model Backbones and Inference Details}
\label{appx:models}

\paragraph{GPT-5.1.}
GPT-5.1 is accessed via the OpenAI API with temperature $T=0$ and default decoding settings.

\paragraph{LLaMA-4-Scout 17B$\times$16E.}
LLaMA-4-Scout 17B$\times$16E is accessed via Together AI in bf16 precision. All four forward passes per query, comprising one original input and three twin worlds, are batched together.

\paragraph{Mistral-Small-24B.}
Mistral-Small-24B is accessed via Together AI in bf16 precision. The same batching strategy is applied as for LLaMA-4-Scout.

\paragraph{Latency.}
Table~\ref{tab:latency-full} reports wall-clock latency for all three backbones under the same query set of 500 instances.

\begin{table}[h]
\centering
\scriptsize
\setlength{\tabcolsep}{6pt}
\renewcommand{\arraystretch}{1.05}
\begin{tabular}{l|ccc}
\toprule
\textbf{Backbone} & \textbf{Passes} & \textbf{Latency (s)} & \textbf{Access} \\
\midrule
GPT-5.1            & 4 & 3.8 $\pm$ 1.3 & OpenAI API \\
LLaMA-4-Scout      & 4 & 3.5 $\pm$ 1.0 & Together AI \\
Mistral-Small-24B  & 4 & 2.9 $\pm$ 1.3 & Together AI \\
\bottomrule
\end{tabular}
\caption{Wall-clock latency per query (mean $\pm$ std) across all three backbones, measured over 500 queries with $k=3$ twin worlds batched together with the original input.}
\label{tab:latency-full}
\end{table}

\subsection{spaCy NER and Delexicalisation}
\label{appx:ner}

We use spaCy's \texttt{en\_core\_web\_trf} pipeline for named entity recognition, mapping its fine-grained labels to four coarse types: \textsc{Person} (\texttt{PERSON}), \textsc{Org} (\texttt{ORG}), \textsc{Location} (\texttt{GPE}, \texttt{LOC}), and \textsc{Date} (\texttt{DATE}). These four types correspond exactly to the pools maintained in the synthetic entity inventory (Appendix~\ref{appx:synthetic-entities}). Entities labelled with other spaCy types (e.g., \texttt{WORK\_OF\_ART}, \texttt{EVENT}) are left unchanged, as the inventory does not cover these types. This conservative choice reduces the scope of substitution but avoids type-mismatched replacements.

Repeated mentions of the same entity are merged using exact string matching followed by alias resolution: a mention is treated as an alias of an earlier entity if it is a strict substring of the canonical form (e.g., \textit{Bale} as an alias of \textit{Gareth Bale}).

\subsection{Human Evaluation Details}
\label{appx:human-eval}

\begin{figure*}[t]
    \centering
    \includegraphics[width=\textwidth]{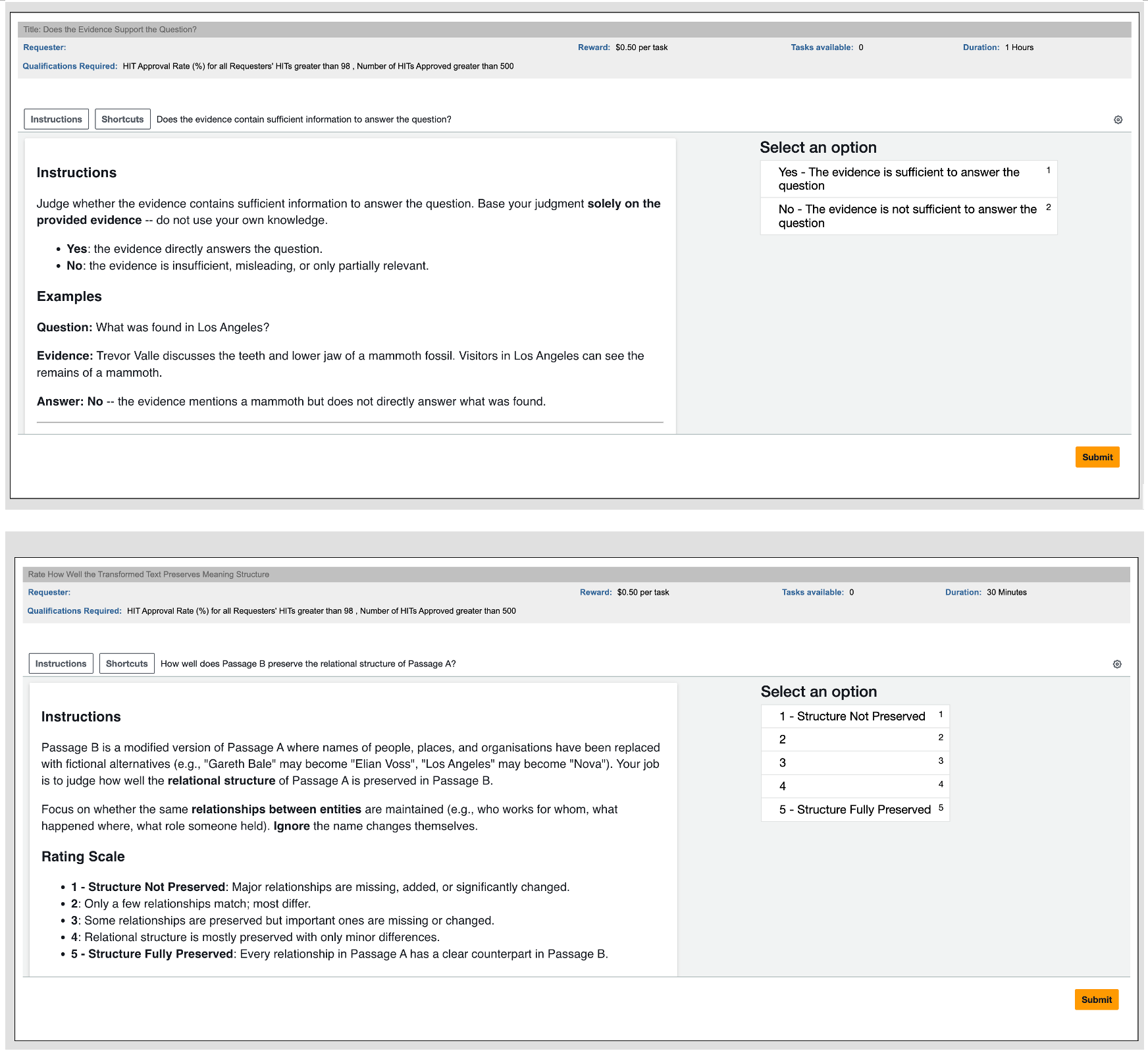}
    \caption{Human-evaluation instructions and annotation interfaces. \textit{Top:} annotators judge whether the evidence is sufficient to answer the question without relying on external knowledge. \textit{Bottom:} annotators rate how well a transformed passage preserves the relational structure of the original passage on a five-point Likert scale.}
    \label{fig:human-eval-instructions}
\end{figure*}

Figure~\ref{fig:human-eval-instructions} shows the annotation interfaces and instructions used in the human evaluation. Annotators complete two tasks. For \textit{abstention alignment}, they judge whether the provided evidence contains sufficient information to answer the question, based solely on the evidence and without using external knowledge. For \textit{structure preservation}, they compare the original and transformed passages and rate preservation of relational structure on a five-point Likert scale, where 1 indicates that major relationships are changed or missing and 5 indicates that all relationships have clear counterparts after transformation.

Each instance is evaluated by three independent annotators. Annotators are restricted to workers with a HIT approval rate above 98\% and more than 500 approved HITs. The structure-preservation task yields a mean rating of 4.61 (SD = 0.48), with 99.1\% of ratings at least 4. For abstention alignment, TW agrees with the majority human judgement on 86\% of answerable instances and 81\% of unanswerable instances, corresponding to 83.5\% overall agreement and Fleiss' $\kappa=0.78$.

\subsection{Agreement Function}
\label{appx:agreement}

We instantiate the agreement function $\mathrm{Agree}(\cdot,\cdot)$ using LLaMA-3-70B as an LLM judge across all four datasets. The judge is prompted with a structured comparison and its output is mapped to $\{0,1\}$. Using a model independent of all three backbone models (GPT-5.1, LLaMA-4-Scout, Mistral-Small-24B) avoids circularity in the evaluation.

For each transformation $\pi$, the agreement function compares the original answer $\hat{y}$ with the back-mapped answer from the corresponding twin world, $\pi^{-1}(\hat{y}^{\,\pi})$, following Section~\ref{sec:methodology}:
\[
S_\pi(q,C)
=
\mathrm{Agree}\!\left(
\pi^{-1}(\hat{y}^{\,\pi}),
\hat{y}
\right).
\]
Gold answers are used only to evaluate final task performance and are not used to compute the equivariance score.

\paragraph{HotpotQA and MIRAGE.}
The judge assesses semantic equivalence between the original answer and the back-mapped twin-world answer, accepting paraphrases, abbreviations, and minor surface variation while rejecting semantically different or incomplete answers. The prompt is:

\begin{llmtext}
Do the following two answers provide the same answer to the question? Answer yes or no only.\\
Question: \{q\}\\
Original answer: \{original\}\\
Back-mapped twin answer: \{twin\}
\end{llmtext}

A ``yes'' judgement is mapped to $1$ and a ``no'' judgement to $0$.

\paragraph{FaithEval.}
The same agreement criterion is used on FaithEval. Since all instances are unanswerable, models may produce abstention-style outputs such as ``no information'' or ``cannot be answered from the evidence.'' The judge assesses whether the original output and back-mapped twin-world output express the same answer behaviour, including semantic equivalence between different abstention formulations.

\paragraph{FEVER.}
For FEVER, generated outputs are normalised to one of three semantic verdicts: yes, no, or neither. The judge uses the following prompt:

\begin{llmtext}
Does the following output indicate that the answer to the claim is yes (supported), no (refuted), or neither? Answer yes, no, or neither only.\\
Claim: \{q\}\\
Output: \{pred\}
\end{llmtext}

Yes is mapped to \textit{Supports}, no to \textit{Refutes}, and neither to \textit{Not Enough Info}. Since these verdicts do not participate in the entity substitution mapping, $\pi^{-1}$ acts as the identity. $S_\pi(q,C)=1$ when the semantic verdict from the twin world agrees with the original verdict and $0$ otherwise. When $s(q,C)<\tau$, TW abstains regardless of the generated verdict and is mapped to \textit{Not Enough Info} at evaluation time.

\subsection{Evaluation Metrics}
\label{appx:metrics}

Following \citet{Kim2025WhenToAbstain, Madhusudhan2025DoLLMsKnow}, we use a confusion matrix (Table~\ref{tab:confusion-matrix}) crossing question types (answerable vs.\ unanswerable) with model behaviours (answered correctly, answered incorrectly, or abstained). TP represents answerable questions answered correctly, FP$_A$ represents answerable questions answered incorrectly, FP$_U$ represents unanswerable questions answered, FN represents answerable questions on which the model abstains, and TN represents unanswerable questions on which the model appropriately abstains.

We report four metrics:
\begin{itemize}[leftmargin=0.5cm]
\item \textbf{F1 Score}: Harmonic mean of precision
$\frac{\mathrm{TP}}{\mathrm{TP}+\mathrm{FP}_A+\mathrm{FP}_U}$
and recall
$\frac{\mathrm{TP}}{\mathrm{TP}+\mathrm{FP}_A+\mathrm{FN}}$.

\item \textbf{Accuracy (Acc)}:
$\frac{\mathrm{TP}+\mathrm{TN}}{N}$,
capturing both correct answering and appropriate abstention.

\item \textbf{Reliability Score (RS)}:
Following \citet{Xu2024Rejection}, we report RS as a complementary measure that jointly considers answer correctness and appropriate abstention. We use the metric definition from the original work without modification.

\item \textbf{Abstention Rate (AR)}:
$\frac{\mathrm{FN}+\mathrm{TN}}{N}$,
measuring the proportion of queries on which the model abstains. Higher AR indicates more conservative behaviour but does not directly imply higher reliability.
\end{itemize}

\begin{table}[h]
\centering
\renewcommand{\arraystretch}{1.5}
\setlength{\tabcolsep}{3pt}
\scriptsize
\begin{tabular}{|c|c|c|c|}
\hline
\multicolumn{2}{|c|}{\multirow{2}{*}{}} 
  & \multicolumn{2}{c|}{Question Type} \\
\cline{3-4}
\multicolumn{2}{|c|}{} 
  & Answerable & Unanswerable \\
\hline
\multirow{2}{*}{Answered} 
  & Correct   & \textcolor{blue}{TP} & \multirow{2}{*}{\textcolor{red}{FP$_U$}} \\\cline{2-3}
  & Incorrect & \textcolor{red}{FP$_A$} & \\
\hline
\multicolumn{2}{|c|}{Abstained}
  & \textcolor{red}{FN} & \textcolor{blue}{TN} \\
\hline
\end{tabular}
\caption{Confusion matrix for abstention evaluation. Blue indicates desirable outcomes; red indicates errors.}
\label{tab:confusion-matrix}
\end{table}

\section{Appendix: Additional Analyses}

\subsection{Parameter Sensitivity Analysis}
\label{appx:parameter-analysis}

\paragraph{Threshold $\tau$.}
We examine sensitivity to the abstention threshold on the held-out validation set. Increasing $\tau$ makes TW more conservative because a larger equivariance score is required to answer, whereas decreasing $\tau$ increases answer coverage. We select $\tau=0.60$ to maximise validation RS. Under the default $k=3$, this corresponds to requiring agreement in at least two of the three twin worlds.

\paragraph{Number of Twin Worlds $k$.}
Table~\ref{tab:k-sweep} reports RS and wall-clock latency per query across values of $k$, averaged across all three backbones. Performance improves steadily from $k=1$ to $k=3$ and changes only marginally thereafter, while computational cost continues to increase. We therefore select $k=3$ as the default.

\begin{table}[t]
\centering
\scriptsize
\setlength{\tabcolsep}{6pt}
\renewcommand{\arraystretch}{1.05}
\begin{tabular}{l|ccc}
\toprule
$k$ & \textbf{Passes} & \textbf{Latency (s)} & \textbf{RS (avg)} \\
\midrule
1 & 2 & 2.1 $\pm$ 0.8 & .910 \\
2 & 3 & 2.9 $\pm$ 0.9 & .942 \\
\textbf{3} & \textbf{4} & \textbf{3.8 $\pm$ 1.3} & \textbf{.954} \\
4 & 5 & 4.6 $\pm$ 1.4 & .955 \\
6 & 7 & 6.3 $\pm$ 1.7 & .956 \\
\bottomrule
\end{tabular}
\caption{RS and latency per query across the number of twin worlds $k$, averaged across all three backbones. Bold indicates the selected default. Gains beyond $k=3$ are negligible relative to the additional inference cost.}
\label{tab:k-sweep}
\end{table}

\subsection{Error Analysis}
\label{appx:error-analysis}

We analyse two sources of error in TW across the full evaluation set of 8,000 instances. Table~\ref{tab:error-analysis} summarises the error rates by type and dataset.

\paragraph{NER Failures.}
spaCy's \texttt{en\_core\_web\_trf} pipeline occasionally misses entity spans or assigns incorrect coarse types, reducing the scope of substitution for affected instances. We identify two failure modes: (i) \textit{missed spans}, where a named entity is not detected at all, and (ii) \textit{type errors}, where an entity is detected but assigned the wrong coarse type, for example a person name classified as \textsc{Org}. Across the full evaluation set, missed spans occur in 4.3\% of instances and type errors in 1.8\%, with higher rates on FEVER (6.1\% and 2.4\%, respectively). Missed spans reduce the number of substituted entities, while type errors may produce less natural substitutions and weaken the resulting equivariance signal.

\paragraph{Boundary Cases.}
With $k=3$, equivariance scores take values in $\{0,0.33,0.67,1.00\}$. We define boundary cases as instances with $s(q,C)\in\{0.33,0.67\}$, the two discrete values surrounding $\tau=0.60$. These account for 8.3\% of the evaluation set. As shown in Figure~\ref{fig:combined-human-eval}, human agreement with TW is lowest in this region, suggesting that equivariance decisions align less strongly with human sufficiency judgements near the decision boundary. Practitioners may adjust $\tau$ according to domain-specific costs of inappropriate answering and abstention.

\begin{table}[t]
\centering
\scriptsize
\setlength{\tabcolsep}{4pt}
\renewcommand{\arraystretch}{1.05}
\begin{tabular}{l|cccc|c}
\toprule
\textbf{Error Type} & \textbf{HotpotQA} & \textbf{MIRAGE} & \textbf{FaithEval} & \textbf{FEVER} & \textbf{Overall} \\
\midrule
Missed spans & 3.8\% & 3.1\% & 4.2\% & 6.1\% & 4.3\% \\
Type errors  & 1.4\% & 1.5\% & 1.9\% & 2.4\% & 1.8\% \\
Boundary     & 7.9\% & 8.1\% & 9.2\% & 8.1\% & 8.3\% \\
\bottomrule
\end{tabular}
\caption{Error rates by type and dataset across the full evaluation set of 8,000 instances. Boundary cases are defined as instances with $s(q,C)\in\{0.33,0.67\}$.}
\label{tab:error-analysis}
\end{table}

\subsection{Computational Cost Analysis}
\label{app:compute-cost}

TW requires $k+1$ forward passes per query: one for the original input and $k=3$ for the twin worlds. The four calls are independent and can be batched or parallelised. Table~\ref{tab:efficiency} reports inference latency and RS on GPT-5.1 over 500 queries.

TW achieves 3.8s mean wall-clock latency with four passes. This is comparable to RC-RAG (3.8s) and Contrastive Decoding (4.1s), while achieving higher RS. At an equal four-pass budget, Self-Consistency trails TW by 0.046 RS. Because entity substitutions approximately preserve input length, token processing scales approximately linearly with the number of worlds; at $k=3$, TW requires approximately four times the token processing of a single-pass inference. Increasing $k$ further produces only negligible RS gains (Table~\ref{tab:k-sweep}).

\begin{table}[t]
\centering
\scriptsize
\setlength{\tabcolsep}{8pt}
\renewcommand{\arraystretch}{1.05}
\begin{tabular}{l|ccc}
\toprule
\textbf{Method} & \textbf{Passes} & \textbf{Latency (s)} & \textbf{RS} \\
\midrule
Zero-shot & 1 & 1.6 $\pm$ 0.6 & .923 \\
Self-Consistency & 4 & 3.6 $\pm$ 1.2 & .910 \\
RC-RAG & 3 & 3.8 $\pm$ 1.4 & .891 \\
Contrastive Decoding & 3 & 4.1 $\pm$ 1.6 & .887 \\
Sufficient Context & 20+ & 17.9 $\pm$ 4.8 & .912 \\
\textbf{TW (Ours)} & 4 & 3.8 $\pm$ 1.3 & \textbf{.956} \\
\bottomrule
\end{tabular}
\caption{Inference latency and RS on GPT-5.1 over 500 queries. TW uses $k+1=4$ forward passes and achieves the highest RS at a latency comparable to other multi-pass baselines.}
\label{tab:efficiency}
\end{table}

\section{Appendix: Case Studies}
\label{appx:case-studies}

\subsection{Case 1: Answerable Question with Faithful Evidence}

\begin{tcolorbox}[colback=gray!10,colframe=black,title=Original Input]
\textbf{Q:} What is \colorbox{green!25}{Matthew McKay}'s occupation? \\
\textbf{E:} \colorbox{green!25}{Canadian} politician \colorbox{green!25}{Matthew McKay} [...] was a \colorbox{green!25}{Liberal} party member of the House of Commons of \colorbox{green!25}{Canada}. He was born in \colorbox{green!25}{West Gwillimbury Township} and became a dentist, dental surgeon and schoolteacher. [...]\\
\textbf{Output:} dentist
\end{tcolorbox}

spaCy detects four entity types in this instance: \textsc{Person} (Matthew McKay), \textsc{Location} (Canada, West Gwillimbury Township), \textsc{Org} (Liberal, House of Commons), and \textsc{Date}. Three structure-preserving transformations $\pi_1,\pi_2,\pi_3$ are sampled, each instantiating a typed bijection over all detected entity spans. The occupation ``dentist'' is not an entity span and therefore does not participate in the substitution mapping; $\pi^{-1}$ acts as the identity on this answer.

\begin{tcolorbox}[colback=blue!5,colframe=blue!50!black,title=Twin World 1]
\textbf{Subst.:} {Matthew McKay} $\mapsto$ {Elian Voss},\; {Canada} $\mapsto$ {Nova},\; {Liberal} $\mapsto$ {Verathen Group} \\
\textbf{Q$^{\pi_1}$:} What is \colorbox{green!25}{Elian Voss}'s occupation? \\
\textbf{E$^{\pi_1}$:} \colorbox{green!25}{Elian Voss} [...] became a dentist, dental surgeon and schoolteacher. [...] \\
\textbf{Output:} dentist \hfill $S_{\pi_1}=1$
\end{tcolorbox}

\begin{tcolorbox}[colback=blue!5,colframe=blue!50!black,title=Twin World 2]
\textbf{Subst.:} {Matthew McKay} $\mapsto$ {Orin Vael},\; {Canada} $\mapsto$ {Valdecor},\; {Liberal} $\mapsto$ {Casen Group} \\
\textbf{Q$^{\pi_2}$:} What is \colorbox{green!25}{Orin Vael}'s occupation? \\
\textbf{E$^{\pi_2}$:} \colorbox{green!25}{Orin Vael} [...] became a dentist, dental surgeon and schoolteacher. [...] \\
\textbf{Output:} dentist \hfill $S_{\pi_2}=1$
\end{tcolorbox}

\begin{tcolorbox}[colback=blue!5,colframe=blue!50!black,title=Twin World 3]
\textbf{Subst.:} {Matthew McKay} $\mapsto$ {Casen Drel},\; {Canada} $\mapsto$ {Mirevan},\; {Liberal} $\mapsto$ {Orven Group} \\
\textbf{Q$^{\pi_3}$:} What is \colorbox{green!25}{Casen Drel}'s occupation? \\
\textbf{E$^{\pi_3}$:} \colorbox{green!25}{Casen Drel} [...] became a dentist, dental surgeon and schoolteacher. [...] \\
\textbf{Output:} dentist \hfill $S_{\pi_3}=1$
\end{tcolorbox}

\noindent\textbf{Decision:}
$s(q,C)=\frac{1}{3}(1+1+1)=1.00\geq\tau=0.60
\Rightarrow \textbf{answer: dentist}$

Across all three twin worlds, the transformed outputs agree with the original answer after back-mapping, yielding $S_{\pi_i}=1$ for all $i$. Because ``dentist'' does not participate in the substitution, equivariance here appears as grounded invariance: the relational predicate remains stable while the surrounding entities change. The consistency across synthetic entity identities provides evidence that the answer is reliably determined by the relational content of the provided evidence.

\subsection{Case 2: Unanswerable Question without Faithful Evidence}

\begin{tcolorbox}[colback=gray!10,colframe=black,title=Original Input]
\textbf{Q:} What was found in \colorbox{green!25}{Los Angeles}? \\
\textbf{E:} [...] \colorbox{blue!20}{Trevor Valle} discusses the teeth and lower jaw of a mammoth fossil. Now, at least 10,000 years later, visitors in \colorbox{green!25}{Los Angeles} can see the remains of a mammoth [...] \\
\textbf{Output:} mammoth fossil
\end{tcolorbox}

spaCy detects two entity types: \textsc{Location} (Los Angeles) and \textsc{Person} (Trevor Valle). The passage mentions a mammoth fossil in relation to Los Angeles but does not provide sufficient evidence to answer the question as posed.

\begin{tcolorbox}[colback=blue!5,colframe=blue!50!black,title=Twin World 1]
\textbf{Subst.:} {Los Angeles} $\mapsto$ {Nova},\; {Trevor Valle} $\mapsto$ {Elian Voss} \\
\textbf{Q$^{\pi_1}$:} What was found in \colorbox{green!25}{Nova}? \\
\textbf{E$^{\pi_1}$:} [...] \colorbox{blue!20}{Elian Voss} discusses the teeth and lower jaw of a mammoth fossil [...] \\
\textbf{Output:} no information \hfill $S_{\pi_1}=0$
\end{tcolorbox}

\begin{tcolorbox}[colback=blue!5,colframe=blue!50!black,title=Twin World 2]
\textbf{Subst.:} {Los Angeles} $\mapsto$ {Valdecor},\; {Trevor Valle} $\mapsto$ {Orin Vael} \\
\textbf{Q$^{\pi_2}$:} What was found in \colorbox{green!25}{Valdecor}? \\
\textbf{E$^{\pi_2}$:} [...] \colorbox{blue!20}{Orin Vael} discusses the teeth and lower jaw of a mammoth fossil [...] \\
\textbf{Output:} no information \hfill $S_{\pi_2}=0$
\end{tcolorbox}

\begin{tcolorbox}[colback=blue!5,colframe=blue!50!black,title=Twin World 3]
\textbf{Subst.:} {Los Angeles} $\mapsto$ {Mirevan},\; {Trevor Valle} $\mapsto$ {Casen Drel} \\
\textbf{Q$^{\pi_3}$:} What was found in \colorbox{green!25}{Mirevan}? \\
\textbf{E$^{\pi_3}$:} [...] \colorbox{blue!20}{Casen Drel} discusses the teeth and lower jaw of a mammoth fossil [...] \\
\textbf{Output:} no answer \hfill $S_{\pi_3}=0$
\end{tcolorbox}

\noindent\textbf{Decision:}
$s(q,C)=\frac{1}{3}(0+0+0)=0.00<\tau=0.60
\Rightarrow \textbf{abstain}$

Across all three twin worlds, the back-mapped outputs fail to agree with the original answer, yielding $s(q,C)=0.00$. The original answer is therefore not stable under structure-preserving entity substitutions and is not reliably determined by the relational structure of the provided evidence. TW consequently abstains.

\end{document}

%% file: main-results.tex
\begin{tabular}{l|cccc|cccc|cccc|cccc} 
\toprule
{} & \multicolumn{4}{c|}{\textbf{HotpotQA}} 
& \multicolumn{4}{c|}{\textbf{MIRAGE}}
& \multicolumn{4}{c|}{\textbf{FaithEval}}
& \multicolumn{4}{c}{\textbf{FEVER}} \\ 
\midrule
 
{Method} 
& {AR} & {F1} & {Acc} & {RS}
& {AR} & {F1} & {Acc} & {RS}
& {AR} & {F1} & {Acc} & {RS}
& {AR} & {F1} & {Acc} & {RS} \\ 
\midrule

& \multicolumn{16}{c}{\raisebox{-.2\height}{\textbf{GPT-5.1}}} \\ 
\midrule
 
{Zero-shot}  
& .0010 & .9325 & \underline{.9320} & \textbf{.9989}
& .0060 & .9157 & .9130 & \underline{.9935}
& .0510 & - & .0510 & .5032
& .3210 & \underline{.6974} & .6260 & \underline{.6620}\\

{AbstentionBench}  
& .0200 & .9232 & .9140 & .9787
& .0580 & .8847 & .8590 & .9372
& .3870 & - & .3870 & .5255
& \textbf{.8430} & .2625 & .4000 & .3618\\

{Self-Consistency}  
& .0110 & \underline{.9346} & .9280 & .9876
& .0180 & \underline{.9276} & \underline{.9250} & .9661
& .1640 & - & .1640 & .5728
& .4280 & .6123 & .5870 & .5904\\

{RC-RAG}  
& \textbf{.2720} & .8090 & .6990 & .7201
& \underline{.2710} & .7912 & .6840 & .7168
& .4860 & - & .4860 & .5004
& .4120 & .4928 & .5400 & .5682\\

{Context Perturbation}  
& .0530 & .9081 & .8840 & .9437
& .0790 & .8631 & .8290 & .9137
& .5090 & - & .5090 & .5002
& .4520 & .5783 & .5030 & .5277\\

{Sufficient Context}  
& \underline{.1430} & .8422 & .7820 & .8463
& \textbf{.4960} & .5944 & .4470 & .4757
& \textbf{.7070} & - & \textbf{.7070} & \textbf{.5857}
& \underline{.4950} & .6360 & .5210 & .5129\\

{Contrastive Decoding}  
& .0170 & .9259 & .9180 & .9819
& .0550 & .8925 & .8680 & .9408
& .3760 & - & .3760 & .5308
& .3940 & .5820 & \textbf{.6690} & .6308\\

\textbf{TW (Ours)}  
& .0030 & \textbf{.9396} & \textbf{.9390} & \underline{.9981}
& .0050 & \textbf{.9318} & \textbf{.9300} & \textbf{.9959}
& \underline{.7010} & - & \underline{.7010} & \underline{.5812}
& .3490 & \textbf{.7215} & \underline{.6550} & \textbf{.6824}\\ 

\midrule
& \multicolumn{16}{c}{\raisebox{-.2\height}{\textbf{LLaMA-4}}} \\ 
\midrule
 
{Zero-shot}  
& .0160 & .8831 & .8760 & \underline{.9823}
& .0100 & .9005 & \underline{.8960} & .9811
& .2470 & - & .2470 & .5280
& .1650 & .6489 & .5740 & .6219\\

{AbstentionBench}  
& .0270 & .8819 & .8700 & .9702
& .0120 & .8964 & .8910 & .9868
& .3100 & - & .3100 & \underline{.5722}
& \textbf{.5020} & .5808 & .5490 & .5236\\

{Self-Consistency}  
& .0210 & \underline{.8849} & \underline{.8790} & .9781
& .0310 & \underline{.9128} & .8890 & .9624
& .4520 & - & .4520 & .5369
& .3710 & .6814 & .6360 & .6128\\

{RC-RAG}  
& \textbf{.3900} & .7106 & .5720 & .5952
& \textbf{.4150} & .7066 & .5600 & .5746
& \underline{.6260} & - & \underline{.6260} & .5318
& \underline{.4190} & .5116 & .5200 & .5554\\

{Context Perturbation}  
& \underline{.1230} & .8556 & .8030 & .8679
& .0450 & .8880 & .8680 & .9511
& .5610 & - & .5610 & .5074
& .4150 & .5691 & .5660 & .5771\\

{Sufficient Context}  
& .1000 & .8663 & .8230 & .8923
& \underline{.1440} & .8384 & .7780 & .8448
& .5830 & - & .5830 & .5138
& \underline{.4190} & \underline{.6977} & \underline{.6540} & .6116\\

{Contrastive Decoding}  
& .0400 & .8816 & .8640 & .9562
& .0110 & .8969 & .8920 & \underline{.9879}
& .3190 & - & .3190 & .5655
& .3750 & .6507 & .6210 & \underline{.6235}\\

\textbf{TW (Ours)}  
& .0130 & \textbf{.8898} & \textbf{.8840} & \textbf{.9869}
& .0090 & \textbf{.9164} & \textbf{.8990} & \textbf{.9916}
& \textbf{.6410} & - & \textbf{.6410} & \textbf{.5791}
& .3860 & \textbf{.7098} & \textbf{.6680} & \textbf{.6327}\\ 

\midrule
& \multicolumn{16}{c}{\raisebox{-.2\height}{\textbf{Mistral-Small}}} \\ 
\midrule

{Zero-shot}  
& .0150 & \underline{.8725} & \underline{.8660} & \underline{.9832}
& .0270 & \underline{.8880} & \underline{.8760} & \textbf{.9704}
& .2880 & - & .2880 & \underline{.5899}
& \underline{.6510} & .4788 & .4950 & .4440\\

{AbstentionBench}  
& .2730 & .7782 & .6720 & .7120
& .1890 & .8172 & .7400 & .7976
& \underline{.6490} & - & \underline{.6490} & .5444
& \textbf{.7050} & .2581 & .3520 & .3352\\
  
{Self-Consistency}  
& .0970 & .8643 & .8110 & .8591
& .1160 & .8782 & .8210 & .8469
& .4720 & - & .4720 & .5518
& .5330 & \underline{.6015} & .5890 & \underline{.5574}\\

{RC-RAG}  
& \textbf{.5370} & .5837 & .4270 & .4437
& \textbf{.4930} & .6397 & .4820 & .4947
& .5580 & - & .5580 & .5067
& .6210 & .5761 & .5700 & .4976\\

{Context Perturbation}  
& .2460 & .7856 & .6890 & .7380
& .2370 & .7760 & .6840 & .7443
& .6230 & - & .6230 & .5303
& .5410 & .5031 & .5330 & .4990\\

{Sufficient Context}  
& .1050 & .8116 & .7690 & .8818
& .2240 & .7815 & .6940 & .7576
& .5710 & - & .5710 & .5101
& .4170 & .4788 & .5500 & \textbf{.5692}\\

{Contrastive Decoding} 
& \underline{.3180} & .7598 & .6390 & .6683
& \underline{.2970} & .7704 & .6560 & .6890
& .6310 & - & .6310 & .5343
& .4590 & .5054 & \underline{.5750} & .5566\\

\textbf{TW (Ours)}  
& .0170 & \textbf{.8798} & \textbf{.8730} & \textbf{.9835}
& .0290 & \textbf{.8942} & \textbf{.8820} & \underline{.9698}
& \textbf{.6560} & - & \textbf{.6560} & \textbf{.5987}
& .4820 & \textbf{.6129} & \textbf{.5860} & .5539\\ 

\bottomrule
\end{tabular}